\documentclass[11pt]{article} 
\usepackage{fullpage,times}

\usepackage{algorithm}
\usepackage{algorithmic}
\usepackage{hyperref}
\usepackage{url}
\usepackage{graphicx} 
\graphicspath{{./figs/}}
\usepackage{subcaption}
\usepackage{censor}
\usepackage{booktabs}
\usepackage{authblk}

\title{Variation Network: Learning High-level Attributes for 
  Controlled Input Manipulation}

\date{}

\author[]{Ga\"etan Hadjeres\thanks{Gaetan.Hadjeres@sony.com} }
\author[]{Frank Nielsen}
\affil[]{Sony Computer Science Laboratories}

\begin{document}

\maketitle

\expandafter\def\expandafter\UrlBreaks\expandafter{\UrlBreaks
  \do\-}

\begin{abstract}
  This paper presents the \emph{Variation Network} (VarNet), a  generative model
  providing means to manipulate the high-level attributes of a given
  input. The originality of our approach is that VarNet is not only capable of
  handling pre-defined attributes but can also learn the relevant
  attributes of the dataset by itself.  These two settings can also be
  easily considered at the same time, which makes this model applicable
  to a wide variety of tasks. Further, VarNet has a sound
  information-theoretic interpretation which grants us with interpretable means to control how these high-level  attributes are learned.
  We demonstrate  experimentally that this model is capable of performing interesting
  input manipulation  and that the learned
  attributes are relevant and meaningful.
\end{abstract}

\section{Introduction}
\label{sec:introduction}
We focus on the problem of learning to generate
\emph{variations} of a given input in an intended way. Concretely, this means that
given some input element $x$, which can be considered as a \emph{template},
we want to generate some transformed versions of $x$ by only modifying its
high-level \emph{attributes}. The objective is that the link between
the original $x$ and its transformed version is preserved while the
difference between their attributes is patent.
Such a mechanism can be of great use in many
domains such as image edition since it allows to edit images using
more abstract controls and can be of crucial importance for creative usages since it
allows to generate new content in a controlled and meaningful way. 


In a  majority of the recent proposed methods tackling this problem such as
\cite{upchurch2016deep,DBLP:journals/corr/LampleZUBDR17,DBLP:journals/corr/abs-1812-04218}, the
attributes of interest that we want to control are assumed to be
given (often given as a discrete variable). 
If these methods are indeed successful in generating meaningful
transformations of any input, we can nonetheless identify two shortcomings which can restrict their use:
1) Labeled data is not always available or can be costly to obtain;
2) Attributes which can be hard to specify in an absolute way cannot be
considered.

The novelty of our approach resides in the fact that it grants the possibility, under
the same framework, to control generations by modifying
user-specified attributes but also to 
\emph{learn} some meaningful high-level attributes at the same time,
which can be used for control.

This problem may seem to be an ill-posed one on many aspects:
Firstly, in the case of specified attributes,
there is no ground truth for variations since there is no $x$ with two
different attributes. Secondly, it can be hard to determine if a
learned attribute is relevant. However, we provide empirical
evidence that our general approach is capable of learning such
relevant attributes and that they can be used for generating
meaningful variations.

This paper introduces the \emph{Variation Network} (VarNet), a
probabilistic neural network which provides means to manipulate an
input by changing its high-level attributes, which can be either
learned or provided.
Our model has a
sound probabilistic interpretation which makes the variations obtained
by changing the attributes statistically meaningful.
This architecture is general and provides a wide range of design
choices.

Our contributions are the following:
\begin{itemize}
\item A widely applicable encoder-decoder architecture which generalizes
  existing approaches
  \cite{2013arXiv1312.6114K,2018arXiv180203761R,DBLP:journals/corr/LampleZUBDR17,DBLP:journals/corr/abs-1812-04218}
  on controlled input manipulation,
\item An easy-to-use framework: any encoder-decoder architecture can be easily
  framed into our framework in order to provide it control over
  generations, even in the case of unlabeled data,
\item Information-theoretic interpretation of our model providing a way to control the behavior of the learned attributes.
\end{itemize}

The plan of this paper is the following: Sect.~\ref{sec:model}
presents the VarNet architecture together with its training
algorithm. For better clarity, we introduce separately all the
components featured in our model and postpone the discussion about
their interplay and the motivation behind our modeling choices in
Sect.~\ref{sec:comments}  We illustrate in
Sect.~\ref{sec:experiments} the possibilities offered by our
proposed model and show that its faculty to generate variations in an
intended way is of particular interest.
Finally, Sect.~\ref{sec:related-work} discusses
about the related works. In particular, we show that VarNet provides
an interesting solution to many constrained generation problems
already considered in the literature while bearing interesting
connections with the literature in fair representations
\cite{2015arXiv151100830L,DBLP:journals/corr/abs-1812-04218} and
disentangled representations
\cite{higgins2016beta,2018arXiv180403599B,locatello2018challenging}.

Our code is available at \url{https://github.com/Ghadjeres/VarNet}.

\section{Proposed model}
\label{sec:model}

We now introduce our novel encoder-decoder architecture which we name
\emph{Variation Network}. Our architecture borrows principles from the
traditional Variational AutoEncoder (VAE) architecture
\cite{2013arXiv1312.6114K} and from the Wasserstein AutoEncoder (WAE)
architecture \cite{2017arXiv171101558T,2018arXiv180203761R}.
It uses an adversarially learned regularization
\cite{2016arXiv160600704D,DBLP:journals/corr/LampleZUBDR17},
introduces a decomposition of the latent space into two parts, a
template $z$ and an \emph{attribute} $\psi$
\cite{pmlr-v80-adel18a} and decomposes the attributes on an adaptive
basis \cite{DBLP:journals/corr/abs-1803-09017}.


We detail in the following sections the different
parts involved in our model. In Sect.~\ref{sec:encdec}, we focus on the encoder-decoder part of
VarNet, in
Sect.~\ref{sec:disentangling}, we introduce the adversarially-learned
regularization whose aim is to disentangle attributes from templates
and in 
Section~\ref{sec:attr-space}, we discuss about the special 
parametrization that we adopted for the space of attributes.


\subsection{Encoder-decoder part}
\label{sec:encdec}
We consider a dataset $\mathcal{D} = \{(x^{(1)}, m^{(1)}), \dots,
(x^{(N)}, m^{(N)})\}$ of $N$ labeled elements $(x,m) \in \mathcal{X} \times
\mathcal{M}$, where $\mathcal{X}$ stands for the input space and
$\mathcal{M}$ for the metadata space. Here, metadata information
is optional as our architecture is also applicable in the totally
unsupervised case. We suppose that the $x$'s follow the data
distribution $\pi(x)$, that for each $x \in \mathcal{X}$ corresponds a unique
label $m$ and write $\pi(x, m)$ the distribution of the $x$'s together
with their metadata information.

Similar to the VAE architectures, we suppose that our data $x \in \mathcal{X}$ depends
on a couple of low-dimensional latent variable $z \in \mathcal{Z} \subset
\mathbf{R}^{d_z}$ and $\psi \in \Psi \subset \mathbf{R}^{d_\psi}$ through some decoder
$p(x|z,\psi)$ parametrized by a neural network. In this paper, we term the $z$
variables \emph{templates} and the $\psi$ variable
\emph{attributes}.  More details about
the attribute space $\Psi$ are given in
Sect.~\ref{sec:attr-space}. 

We introduce a factorized prior $p(z, \psi) = p(z)p(\psi)$
over this latent space so that the joint probability distribution is
expressed as
$p(x,z,\psi) = p(x|z,\psi)p(z)p(\psi)$. The objective is to maximize
the log-likelihood $\mathbf{E}_{(x,m) \sim \mathcal{D}} \log \pi(x)$ of
the data under the model.
Since the posterior distribution $p(z|x)$ is
usually intractable, an approximate posterior distribution $q(z|x)$
parametrized by a neural network is
introduced (for simplicity, we let the approximate posterior depend on
$x$ only, but considering $q(z|x,m)$ or even $q(z|x,\psi)$ is feasible). Concerning the computation of $\psi$ given $x$, we introduce a
\emph{deterministic} function $f(x,m)$ parametrized by a neural network, which can eventually rely on metadata
information $m$.


We then obtain  the
following mean reconstruction loss:
\begin{equation}
  \label{eq:re}
\mathrm{RE} :=  -\mathbf{E}_{(x,m) \sim \pi} \mathbf{E}_{z
  \sim q(z| x)}  \log p(x | z, f(x,m)),
\end{equation}

and regularize the latent space  $\mathcal{Z}$ by
adding  the usual Kullback-Leibler ($\mathrm{KL}$) divergence  term  appearing in the VAE Evidence
Lower Bound (ELBO) on $\mathcal{Z}$:
\begin{equation}
  \label{eq:kl}
\mathrm{KL} := \mathbf{E}_{z \sim q(z|x)} \log \frac{q(z|x)}{p(z)}.
\end{equation}


\subsection{Disentangling attributes from templates}
\label{sec:disentangling}

Our decoder $p(x| z, \psi)$ thus depends exclusively on $z$ and on
features $\psi$.
However, there is no reason for a random
attribute $\psi \in \Psi \neq f(x, m)$ that $\hat{x} \sim p(.|z,\psi)$, where $z
\sim q(z|x)$, is a meaningful
variation of the original $x$. Indeed, all needed information for
reconstructing $x$ could
potentially be already contained in $z$ and changing $\psi$ could have
no effect on the reconstructions. 

To enforce this property, we propose to add an adversarially-learned cost on the latent
variables $z$ and $\psi$ to force
the encoder $q$ to discard information about the attribute $\psi = f(x,m)$ of
$x$: Specifically, we train a discriminator neural network
$D:\mathcal{Z} \times \Psi \to [0, 1]$ whose role is to evaluate the
probability $D(z, \psi)$ that there exists a pair $(x, m) \in
\mathcal{D}$ such that $\psi = f(x, m)$ and $z \sim q(z|x)$. In
other words, the aim of the discriminator is to determine if the attributes
$\psi$ and the template code $z$ originate from the same $(x, m) \in
\mathcal{D}$ or if the features $\psi$ are sampled from the prior
$p(\psi)$.


The encoder-decoder architecture
presented in Sect.~\ref{sec:encdec} is trained to fool the
discriminator: this means that, for a given $(x, m) \in \mathcal{D}$, it tries to  produce a template code
$z \sim q(z|x)$ and features $\psi = f(x,m)$ such that no information about
the features could be recovered from $z$.

In an optimal setting, i.e. when the discriminator is
unable to match any $z \in \mathcal{Z}$ with a particular feature $\psi \in
\Psi$,  the space of template codes and the space of
attributes are decorrelated and the aggregated distribution of the
$\psi$'s $f_*(\pi)$ (the pushforward measure of $\pi$ by $f$) 
matches the prior $p(\psi)$.


The discriminator is trained to maximize

\begin{equation}
  \label{eq:disctrain}
\mathcal{L}_\mathrm{Disc} := \mathbf{E}_{(x,m) \sim \pi} \mathbf{E}_{z \sim q(z|x)}  \mathbf{E}_{\psi \sim p(\psi)} \big[\log
  D(z, \psi) + \log(1 - D(z, f(x, m)))
\big].
\end{equation}

while the encoder-decoder architecture is trained to minimize
\begin{equation}
  \label{eq:4}
  \mathcal{R}_\mathrm{Disc} := - \mathbf{E}_{(x,m) \sim \pi}
  \mathbf{E}_{z \sim q(z|x)}  \mathbf{E}_{\psi \sim p(\psi)} \big[ \log  D(z, f(x, m)) + \log( 1 - D(z,\psi)) \big].
\end{equation}

This is the same setting as in
\cite{2016arXiv160600704D,2017arXiv171101558T} but other GAN training
methods could be considered
\cite{DBLP:journals/corr/abs-1902-05687}. Note  the presence of two terms in
Eq.~(\ref{eq:4}), since the ``true'' examples depend on training
parameters.

The final objective we minimize for the encoder-decoder architecture is a combination of a reconstruction
loss Eq.~(\ref{eq:re}), a Kullback-Leibler penalty on $z$
Eq.~(\ref{eq:kl}) and the adversarially-learned loss
Eq.~(\ref{eq:4}). 

\begin{figure}[t]
\centering
\includegraphics[width=0.4\textwidth]{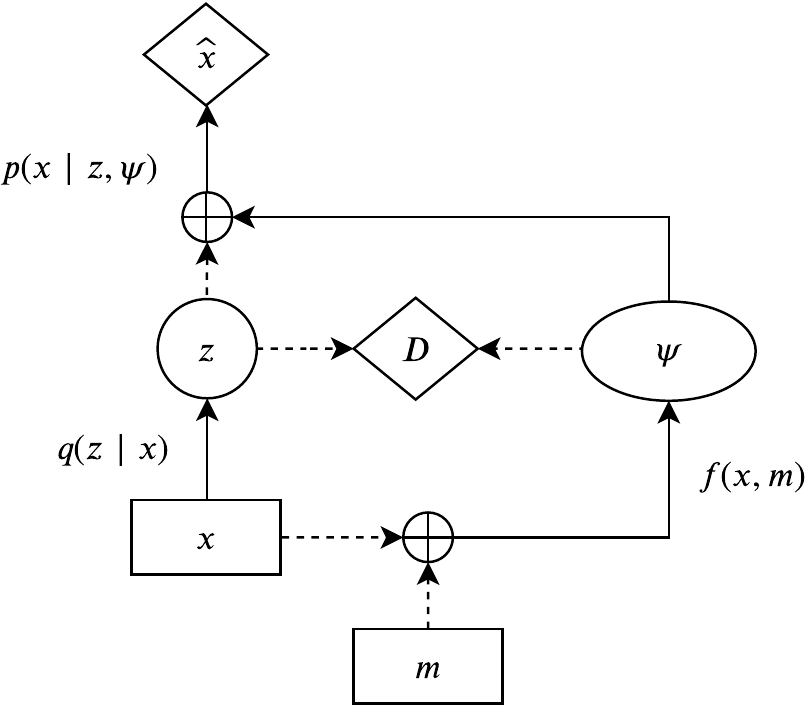}
\caption{VarNet architecture. The input $x$, and its reconstruction
  $\hat{x}$ are in the input space $\mathcal{X}$, the metadata $m$ is
  in the metadata space  $\mathcal{M}$, the latent variable $z$ lies in
  $\mathcal{Z}$ the latent template
  space and the attributes $\psi$ are in the attribute space $\Psi$.
  Dotted arrows represent identity mappings, filled arrows represent
  neural networks and the circled plus sign represents the
  concatenation of its two incoming arguments.
The
discriminator $D$ acts on $\mathcal{Z} \times \Psi$.
}
\label{fig:model}
\end{figure}

Our final architecture is shown in
Fig.~\ref{fig:model} and the proposed training procedure in shown in
Alg.\ref{alg:1}. Estimators of Eq.~(\ref{eq:disctrain}), Eq.~(\ref{eq:re}) and (\ref{eq:4}) are given by
Eq.~(\ref{eq:3}), Eq.~(\ref{eq:6}) and (\ref{eq:rdisc}) respectively. The
the estimator $\mathrm{KL}_n$ of the KL term can be either sampled or computed in closed form.

\begin{algorithm}[h!]
  \caption{Variation Network training procedure}
  \label{alg:1}
  \begin{algorithmic}[1]
    \REQUIRE Dataset $\mathcal{D} = \left\{(x^{(i)},
      m^{(i)}\right\}_{i=1..N}$, reconstruction cost $c$,\\reproducing kernel $k$, batch size
    $n$
    \FOR{Fixed number of iterations}
    \STATE Sample $x := (x_1, \dots,  x_n)$ and $m := (m_1, \dots, m_n)$
    where $(x_i, m_i)$ are i.i.d. samples from $\mathcal{D}$
    \STATE Sample $z_i \sim q(z|x_i)$
    
    \STATE Sample $\hat{x} := \{ \hat{x}_1, \dots, \hat{x}_n \}$ where $\hat{x}_i \sim p(x|z_i,f(x_i,m_i)$,
    \STATE Sample random features $\{\psi_i\}_{i = 1..n} \sim p(\psi)$ 

    \STATE \emph{Discriminator training phase}
    \STATE Compute
    \begin{equation}
      \label{eq:3}
\mathcal{L}_{\textrm{Disc}, n} := \frac{1}{n}\sum_{i=1}^n \log D\left(z_i, \psi_i\right)
      + \log \left(1 - D(z_i, f(x_i, m_i))\right)
    \end{equation}

    \STATE Gradient ascent step on the discriminator parameters using
    $\nabla \mathcal{L_\textrm{Disc}}$
    \STATE \emph{Encoder-decoder training phase}
    \STATE Compute
    \begin{equation}
      \label{eq:lenc}
\mathcal{L}_{\mathrm{EncDec}, n} := \mathrm{RE}_n + \beta
\mathrm{KL}_n  + \gamma  \mathcal{R}_{\mathrm{Disc}, n}
    \end{equation} where
    \begin{equation}
      \label{eq:6}
      \mathrm{RE}_n := - \frac{1}{n} \sum_{i=1}^n \log p (x_i|z_i,f(x_i,m_i)),
    \end{equation}

     \begin{equation}
  \mathcal{R}_{\mathrm{Disc}, n} = - \frac{1}{n}\sum_{i=1}^n 
    \log  D(z_i, f(x_i, m_i))
    + \log (1 -  D(z_i, \psi_i)).
\label{eq:rdisc}
\end{equation}
    \STATE Gradient ascent step on all parameters except on the discriminator parameters using
    $\nabla \mathcal{L_\textrm{EncDec}}$
    \ENDFOR
  \end{algorithmic}
\end{algorithm}

\subsection{Parametrization of the attribute space}
\label{sec:attr-space}
We adopt a particular parametrization of our attribute function
$f: \mathcal{X} \times \mathcal{M}$. In the following, we make a distinction between
two different cases: the case of
continuous \emph{free attributes} and the case of fixed continuous or
discrete attributes.

\subsubsection{Free attributes}
\label{sec:free-attr}
In order to handle free attributes, which denote attributes that are
not specified \emph{a priori} but learned  we introduce $d$ \emph{attribute vectors} $v_i$ of dimension $d_\psi$ together with an
\emph{attention module} $\alpha: \mathcal{X} \times \mathcal{M} \to [0,
1]^{d}$.  By denoting $\alpha_i$ the coordinates of $\alpha$, we
then write our attribute function $f$ as
\begin{equation}
  \label{eq:2}
 f(x, m) = \sum_{i=1}^{d} \alpha_i(x, m) v_i. 
\end{equation}

This approach is similar to the \emph{style tokens} approach presented
in \cite{DBLP:journals/corr/abs-1803-09017}. The $v_i$'s are global
and do not depend on a particular instance $(x, m)$.  By varying the values of the
$\alpha_i$'s between $[0, 1]$, we can then span a $d$-dimensional
hypercube in $\mathbf{R}^{d_\psi}$ which stands for our
\emph{attribute space} $\Psi$. It is worth noting that the $v_i$'s are
also learned and thus constitute an adaptive basis of the attribute
space.

The prior $p(\psi)$ over $\Psi$ (note that
this subspace also varies during training) is obtained by pushing forward fixed
distribution $\nu_\alpha$ over $[0, 1]^{d}$ (often considered to be
the uniform distribution) using Eq.~(\ref{eq:sampling-psi}):
\begin{equation}
  \label{eq:sampling-psi}
  \psi \sim p(\psi) \quad \Longleftrightarrow \quad \psi =
  \sum_{i=1}^{d}\alpha_i v_i \quad \textrm{where} \quad \alpha_i \sim \nu_\alpha.
\end{equation}

\subsubsection{Fixed attributes}
\label{sec:fixed-attr}
We now suppose that the metadata variable $m \in \mathcal{M}$ contains
 attributes that we want to vary at generation time. For
simplicity, we can suppose that this metadata information can be
either continuous  with values in $[0, 1]^M$ (with a natural order on
each dimension) or discrete with values in
$[|0, M|]$. 

In the continuous case, we write our attribute function
\begin{equation}
  \label{eq:cont-attr}
  f(x, m) = \sum_{i=1}^M m_i v_i
\end{equation}
while in the discrete case, we just consider
\begin{equation}
  \label{eq:disc-attr}
  f(x, m) = e_m,
\end{equation}
where $e_m$ is a $d_\psi$-dimensional embedding of the symbol $m$. It
is important to note that even if the attributes are fixed, the
$v_i$'s or the embeddings $e_m$ are learned during training.

These two equations define a prior $p(\psi)$ via:
\begin{equation}
 \psi \sim p(\psi) \quad \Longleftrightarrow \quad \psi = f(x, m) \quad
  \textrm{where} \quad (x, m) \sim \mathcal{D}.\label{eq:nu-fixed}
\end{equation}


\section{Comments}
\label{sec:comments}
We now detail our objective (\ref{eq:lenc}) and notably
explicit connections in Sect.~\ref{sec:infotheory} with \cite{DBLP:journals/corr/abs-1812-04218}
which gives an information-theoretic interpretation of a similar
objective.
In Sect.~\ref{sec:attr-fun}, we further discuss about
the multiple possibilities that we have concerning the implementation of the
attribute function and list, in Sect.~\ref{sec:sampling-schemes},
the different sampling schemes of VarNet. 

\subsection{Information-theoretic interpretation}
\label{sec:infotheory}
Our objective bears similarity the objective Eq.~(11) from
\cite{DBLP:journals/corr/abs-1812-04218} which is derived from the
following optimization problem
\begin{equation}
  \label{eq:infoopt}
  \max I(x;z|\psi) \quad \textrm{with} \quad I(z;\psi) < \epsilon,
\end{equation}
where $I(\cdot;\cdot)$ and $I(\cdot;\cdot|\cdot)$ denote the mutual
information and the conditional mutual information respectively (using
the joint distribution $q(x,z,f(x,m)) = \pi(x,m) q(z|x)$).

The differences between our work and
\cite{DBLP:journals/corr/abs-1812-04218} are the following:
First, our objective Eq.~(\ref{eq:lenc}) is motivated from a
generative point of view, while the motivation in
\cite{DBLP:journals/corr/abs-1812-04218} is to obtain fair
representations
(\cite{2015arXiv151100830L,xie2017controllable,moyer2018invariant}). This
allows us to consider attributes computed from both $x$ and $m$, while
in \cite{DBLP:journals/corr/abs-1812-04218} we have $\psi = f(x,m) =
m$ and the ``sensitive attributes'' $m$ must be provided.
A second difference is our formulation of the discriminator
function: \cite{DBLP:journals/corr/abs-1812-04218} considers training a discriminator to
predict $\psi$ given $z$. In our case, such a formulation would
prevent us to impose a prior distribution over $\Psi$.

The connection with the work of
\cite{DBLP:journals/corr/abs-1812-04218} thus provides an
interpretation of hyperparameters $\beta$ and $\gamma$ in Eq.~(\ref{eq:lenc}).

\subsection{Flexibility in the choice of the attribute function}
\label{sec:attr-fun}
In this section, we focus on the parametrization of the attribute
function $f: \mathcal{X} \times \mathcal{Z} \mapsto
\mathbf{R}^{d} $. 
The formulation of Sect.~\ref{sec:attr-space} is in fact too restrictive
and considered only particular attribute functions. It is in fact
possible to mix different attribute functions by simply concatenating
the resulting vectors. By doing so, we can then
combine free and fixed attributes in a natural way but also consider
different attention modules $\alpha$
similarly to what is done in
\cite{DBLP:journals/corr/ChenKSDDSSA16}, but also consider different
distributions over the attention vectors $\alpha_i$.

It is important to note that the free attributes presented in
Sect.~\ref{sec:free-attr} can only capture \emph{global} attributes,
which are attributes that are relevant for \emph{all} elements of the
dataset $\mathcal{D}$. In the presence of discrete labels $m$, it can
be interesting to consider \emph{label-dependent free attributes},
which are attributes specific to a subset of the dataset. In this
case, the attribute function $f$ can be written as
\begin{equation}
  \label{eq:label-dep-attr}
  f(x, m) = \sum_{i=1}^{d} \alpha_i(x, m) e_{m, i},
\end{equation}
where $e_{m, i}$ designates the $i^{th}$ attribute vector of the label
$m$.
With all these possibilities at hand, it is possible to devise numerous
applications in which the notions of \emph{template} and
\emph{attribute} of an input $x$ may have diverse interpretations.

Our choice of using a discriminator over $\Psi$ instead of, for
instance, over the values of $\alpha$ themselves allow to encompass
within the same framework discrete and continuous fixed
attributes. This also makes the combinations of such attribute functions natural.

\subsection{Sampling schemes}
\label{sec:sampling-schemes}
We quickly review the different sampling schemes of VarNet. We believe
that this wide range of usages makes VarNet a promising model for a
wide range of applications. We can for instance:
\begin{itemize}
\item generate random samples $\hat{x}$ from the estimated dataset
  distribution:
  \begin{equation}
    \label{eq:1}
\hat{x} \sim p(x|z, \psi) \quad \textrm{with} \quad  z \sim p(z) \quad \textrm{and} \quad
\psi \sim p(\psi),
  \end{equation}
\item sample $\hat{x}$ with given attributes $\psi$:
  \begin{equation}
    \label{eq:5}
\hat{x} \sim p(x|z, \psi) \quad \textrm{with} \quad z \sim p(z),
  \end{equation}
\item generate variations of an input $x$ with attributes $\psi$:
  \begin{equation}
    \label{eq:8}
\hat{x} \sim p(x|z,\psi) \quad \textrm{with}  \quad z \sim q(z|x),
  \end{equation}
\item generate random variations of an input $x$:
  \begin{equation}
\hat{x} \sim p(x|z,\psi) \quad \textrm{with}  \quad z \sim q(z|x) \quad \textrm{and} \quad
\psi \sim p(\psi).\label{eq:9}
\end{equation}
\end{itemize}

In the case of continuous attributes of the form Eq.~(\ref{eq:2}) or
(\ref{eq:cont-attr}), VarNet also provides a new way to generate
interpolations, since we can perform interpolations either on $z$ or
on $\psi$.

\section{Experiments}
\label{sec:experiments}
\label{sec:mnist}
We now illustrate the different sampling schemes presented in
Sect.~\ref{sec:sampling-schemes} on different image
datasets. Section~\ref{sec:impl-details} is devoted to the
implementation details and our particular modeling choices,
Sect.~\ref{sec:generation-examples} and \ref{sec:underst-free-attr} showcase some applications of VarNet
ensuring that the learned attributes are meaningful and controllable.
Finally, Sect.~\ref{sec:inductive-biases} discusses the impact of the modeling choices.

\subsection{Implementation details}
\label{sec:impl-details}
In all these experiments, we choose to use a ResNet-50
\cite{he2016deep} for the encoder network. We also use a ResNet-50 for the attribute
function $f$ when the function $f$ depends on the input $x$ and not only on
metadata information $m$. For the decoder network, we use a
conditional transposed
ResNet-50 similar to the decoder used in
\cite{DBLP:journals/corr/abs-1806-01054}. The batch normalization
parameters \cite{ioffe2015batch} are conditioned by $\psi$ similarly
to what is done in \cite{dumoulin2018feature,perez2018film}.

Following \cite{dai2019diagnosing}, we consider a simple parametrization
for the encoder $q(z|x)$ and decoder $p(x|z,\psi)$ networks: the encoder $q(z|x)$ is a
diagonal Gaussian distribution
$\mathcal{N}(\mu_z(x), \textrm{diag}(\sigma_z(x)))$
and the decoder $p(x|z,\psi)$ is a diagonal Gaussian distribution with
scalar covariance matrix $\mathcal{N}(\mu_x(z,\psi), \sigma_x I)$, where $\sigma_x$ is a
learnable scalar parameter and where $I$ denotes the identity matrix.

\label{sec:results-sampl}

The prior $p(z)$ over $\mathcal{Z}$ is a unit centered Normal distribution.
For the sampling of the $\alpha$ values in the free attributes case, we considered $\nu_\alpha$ to
be a uniform
distribution over $[0,1]^{d}$. In the fixed attribute case, we
simply obtain a random sample $\{\psi_i\}_{i=1}^n$ by shuffling the
already computed batches of $\{f(x_i, m_i)\}_{i=1}^n$ (lines 4 and
6 in Alg.\ref{alg:1}).

The dimension of the template space $\mathcal{Z}$ is $d_z=16$ for
applications on the
MNIST \cite{lecun1998mnist}, KMNIST \cite{clanuwat2018deep},
Fashion-MNIST \cite{xiao2017/online} and
dSprites \cite{higgins2016beta,dsprites17} datasets; and is equal to
$d_z=64$ on the CelebA dataset \cite{liu2015faceattributes} with
resolution $64 \times 64$. The hyperparameter $\beta$ and $\gamma$ in
Eq.~(\ref{eq:lenc}) are chosen equal
to $1$ and $10$ respectively.

\subsection{Influence of the attribute function}
\label{sec:generation-examples}
We now apply VarNet on the different image datasets mentioned above in order to illustrate the different
sampling schemes presented in Sect.~\ref{sec:mnist} and the
influence of  the choice of the 
attribute functions (Sect.~\ref{sec:attr-fun}). We primarily focus on
sampling schemes Eq.~(\ref{eq:8}) and Eq.~(\ref{eq:9}) which help
understanding the effect of the learned attributes.

\subsubsection{Fixed discrete attributes}
\label{sec:fixattr}
\begin{figure}[h!]
  \center
  \subcaptionbox{  \label{fig:mnistlabel}}{\includegraphics[width=2.8in]{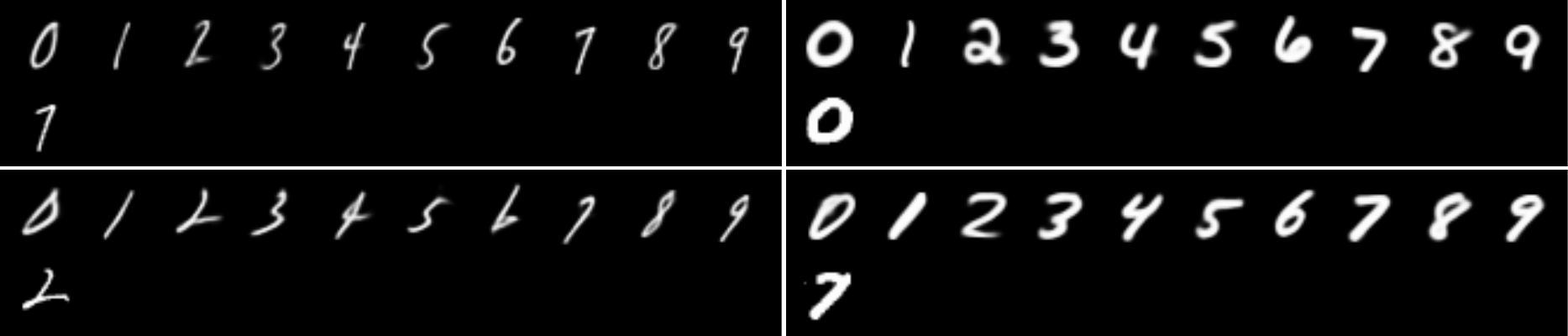}}
  \quad
  \subcaptionbox{  \label{fig:1f1lvars2}}{\includegraphics[width=2.8in]{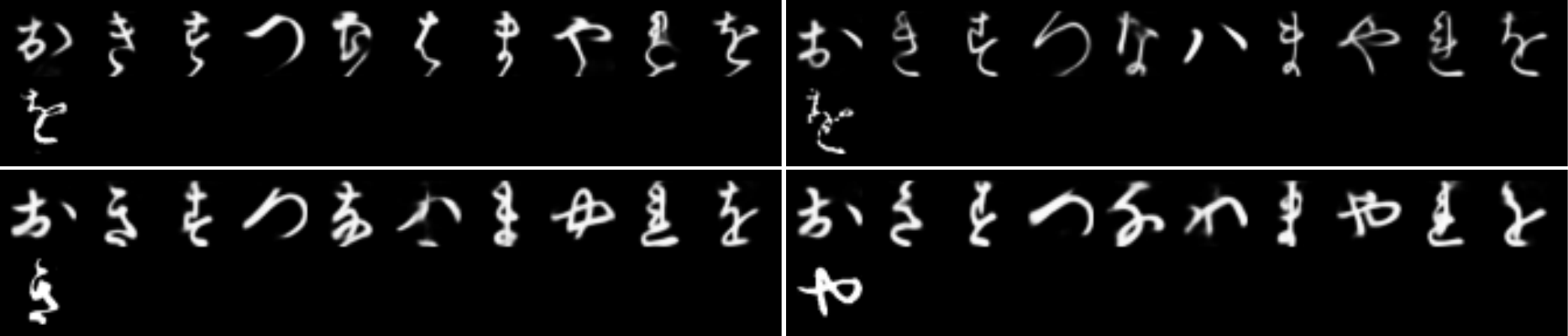}}
  \caption{Visualization of the spanned \emph{space of variations} for
    an attribute function on MNIST (a)
    and KMNIST (b). The original input $x$ is shown on the bottom left of
    each image and the reconstructions for all 10 labels are shown on
    the upper line.}
 \label{fig:varslabel}
\end{figure}

We start by considering attribute functions $f$ depending exclusively
on discrete labels $m$
(Eq.~(\ref{eq:disc-attr})). Figure~\ref{fig:varslabel} shows the
effect of changing the attributes of a given input $x$ using
Eq.~(\ref{eq:8}) on the MNIST and KMNIST datasets. We observe consistency between all variations. In
these two cases, the template $z$ encompasses the handwriting and it is
interesting noting that characters with two distinct way of writing appear
in the generated variations.

\subsubsection{Free continuous attributes}
\label{sec:free-attrs}
\begin{figure}[h!]
  \subcaptionbox{  \label{fig:mnistfree}}{\includegraphics[width=1.9in]{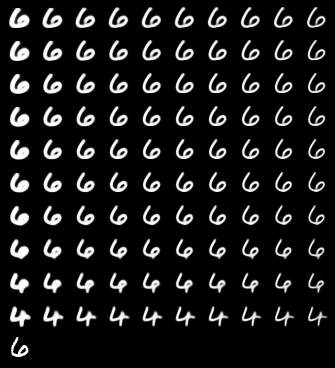}}
  \hfill
    \subcaptionbox{  \label{fig:dspritesfree}}{\includegraphics[width=1.9in]{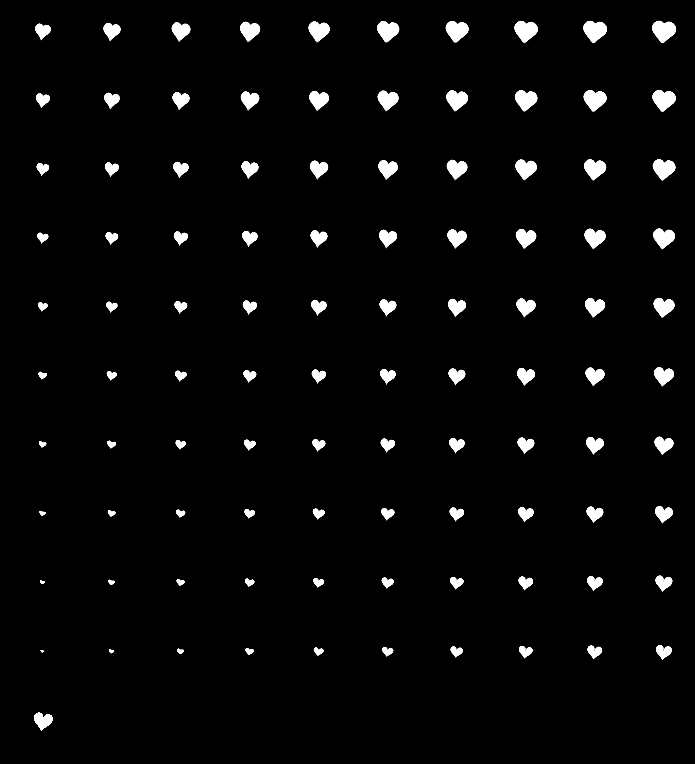}}
   \hfill
  \subcaptionbox{  \label{fig:celebafree}}{\includegraphics[width=1.9in]{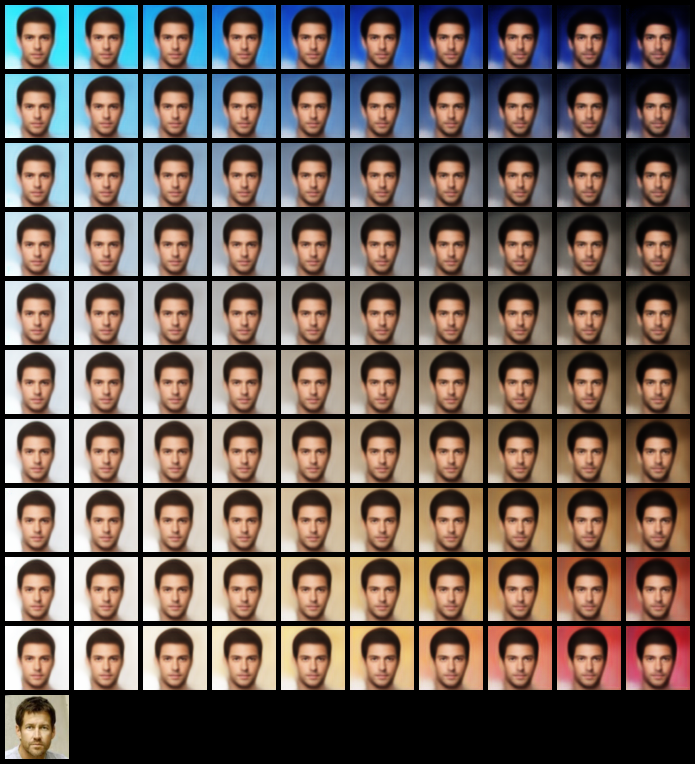}}
  \caption{Visualization of the spanned \emph{space of variations}
    obtained by
    varying uniformly a 2-dimensional learned free attribute on
    different datasets: (a) MNIST (b) dSprites (c) CelebA.
  The initial input $x$ is displayed on the bottom left of each plot.}
 \label{fig:varsfree}
\end{figure}

We now focus on the case where we let the network learn relevant
attributes of the data. Figure~\ref{fig:varsfree} displays typical
examples of variations obtained by considering free attributes of
dimension $d=2$ on different datasets.  These plots are obtained by
varying the $\alpha_i$ in Eq.~(\ref{eq:sampling-psi}) at a constant
speed between 0 and 1.

In all cases, the learned attributes are relevant to the dataset and
meaningful. 
We find interesting to note that there is no disentanglement
\emph{between} the dimensions of the attributes and that the learned
attributes are highly dependent on the dataset at hand.

\subsubsection{Mixing attributes}
In this part, we give examples of attribute functions  obtained
by combining  the ones mentioned in the preceding sections. In
particular, we consider attribute functions created by combining
\begin{itemize}
\item  several discrete fixed attributes,
\item  discrete fixed attributes and continuous free attributes,
\item  continuous free attributes conditioned by discrete fixed
  attributes.
\end{itemize}
The objective is to showcase how the design of the
attribute functions influences the learned attributes and the ``meaning''
of the templates $z$.

\begin{figure}[h!]
  \includegraphics[width=1.6in]{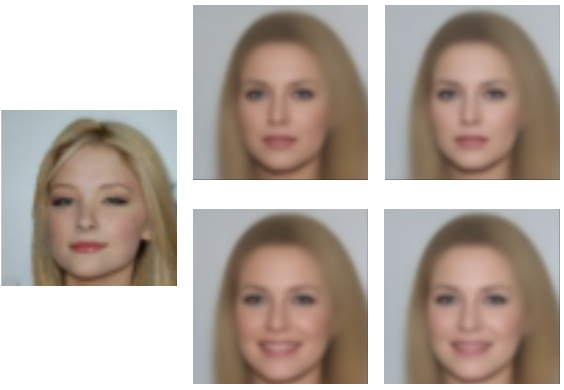}
  \centering
  \caption{Visualization of the spanned \emph{space of variations} for
    an attribute function depending on 2 binary labels, ``Smiling''
    (in lines)
    and ``Pale Skin'' (in columns), on CelebA. The original input is
    shown on the left.
}
\label{fig:celebalabel}
\end{figure}
Figure~\ref{fig:celebalabel} considers the case where two attribute
functions relying solely on binary labels are concatenated. We observe
that the resulting model allow to vary independently every label
during generation.
Such a design choice can be useful since concatenating two attributes
functions taking into account two different labels instead of choosing a unique attribute function
taking into account a Cartesian product of labels tends to make
generalization  easier.

We now consider the case where the free attributes are combined with
fixed attributes. There are two different ways to do so: one is to
concatenate a attribute function on free attributes with one taking
into account fixed discrete attributes, the other is to make the free
attributes \emph{label-dependent} as in Eq.~(\ref{eq:label-dep-attr}).

\begin{figure}[h!]
  \centering
  \subcaptionbox{\label{fig:freefixed}}{
    \includegraphics[width=3.4in]{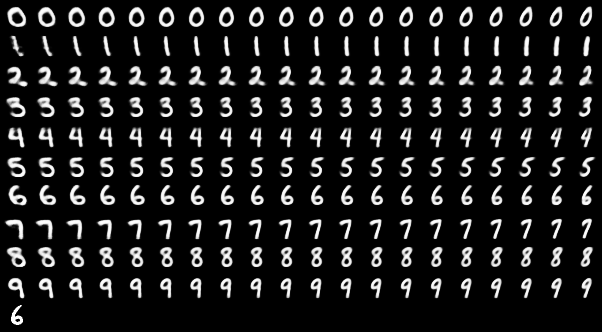} \vspace{.2in}}
  \quad \hfill
  \subcaptionbox{\label{fig:addidas}}{\includegraphics[width=2.1in]{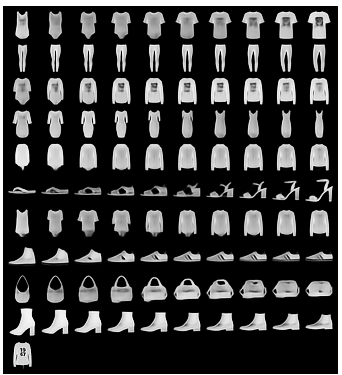}}
  \caption{Visualization of the spanned \emph{space of variations}:
    (a) for
    attributes being the concatenation of a fixed discrete attribute
    and a 1-dimensional free attribute on MNIST, (b) for
    a 1-dimensional label-dependent attribute
    Eq.\ref{eq:label-dep-attr}) on Fashion-MNIST.}
\end{figure}

Figure~\ref{fig:freefixed} displays one such example in the first
case. This corresponds to an attribute function $f$ as in
Eq.~(\ref{eq:label-dep-attr}), but where the dependence of $\alpha_i$ on
$m$ is dropped. This results in the fact that the learned free
attributes are global and possess the same ``meaning'' for all
classes. In Fig.~\ref{fig:addidas}, on the contrary, the dependence of
the $\alpha_i$'s on $m$ is kept. This has the effect that each class
possesses a ``local'' free attribute.

\subsection{Influence of the hyperparameters}
\label{sec:underst-free-attr}

From the preceding examples of Sect.~\ref{sec:generation-examples}, we saw that the fixed label
attributes have clearly been taken into account, but it can be hard to
guess a priori which
high-level attribute  the free attribute function might
capture. However, for a given architecture, a given dataset and fixed hyperparameters,
we observed that the network tended to learn the same high-level
features across multiple runs. In this section, we show that the
information-theoretic interpretation from Sect.~\ref{sec:infotheory}
gives us an additional way to control what is learned by the free
attributes by modifying the hyperparameter $\beta$ and/or $\gamma$.

\begin{figure}[h!]
  \centering
  \subcaptionbox{\label{fig:vars2D1}}{\includegraphics[width=0.3\textwidth]{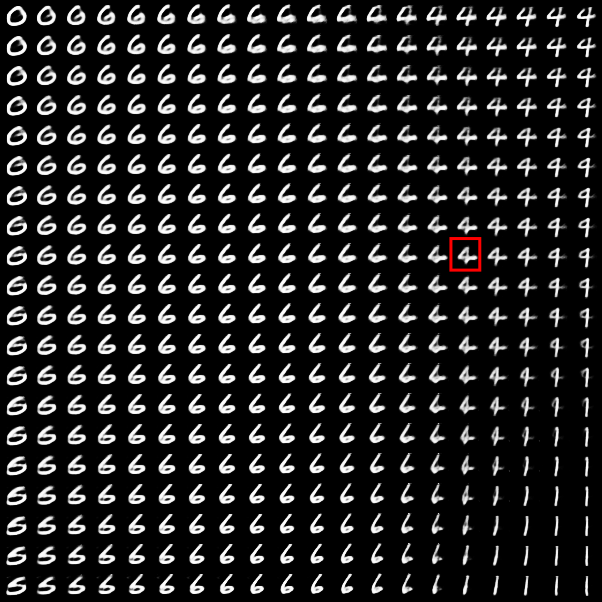}}
  \quad 
  \subcaptionbox{\label{fig:lowKL}}{\includegraphics[width=0.3\textwidth]{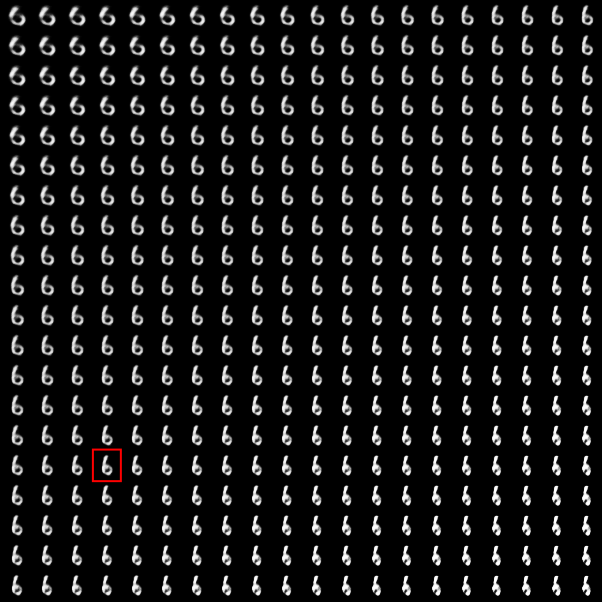}}

  \caption{Figures \ref{fig:vars2D1} and \ref{fig:lowKL} display the
    space of variations using a 2-dimensional free attribute on MNIST. Figure~\ref{fig:lowKL} was generated using a model
    trained with a low KL penalty ($\beta=0.1$). The closest
    generation from the input lies in the red square.}
  \label{fig:vars2d}
\end{figure}

For some applications, variation spaces such as the one displayed in
Fig.~\ref{fig:vars2D1} are not
desirable because they may tend to move too ``far away'' from  the
original input. As discussed in Sect.~\ref{sec:infotheory}, it is possible to reduce how ``spread''
the spaces of variation are by modifying the $\beta$ parameter multiplying the KL
term in the objective Eq.~(\ref{eq:lenc}). An example of such a variation space is
displayed in Fig.~\ref{fig:lowKL}. If the interpretation of the
high-level features 
learned by the free attributes seems identical in both cases, the
variations spanned in \ref{fig:lowKL} are all more similar from the
original input.

\subsection{Inductive biases}
\label{sec:inductive-biases}
\begin{figure}[h!]
  \centering
  \subcaptionbox{\label{fig:celeba1}}{\includegraphics[width=0.3\textwidth]{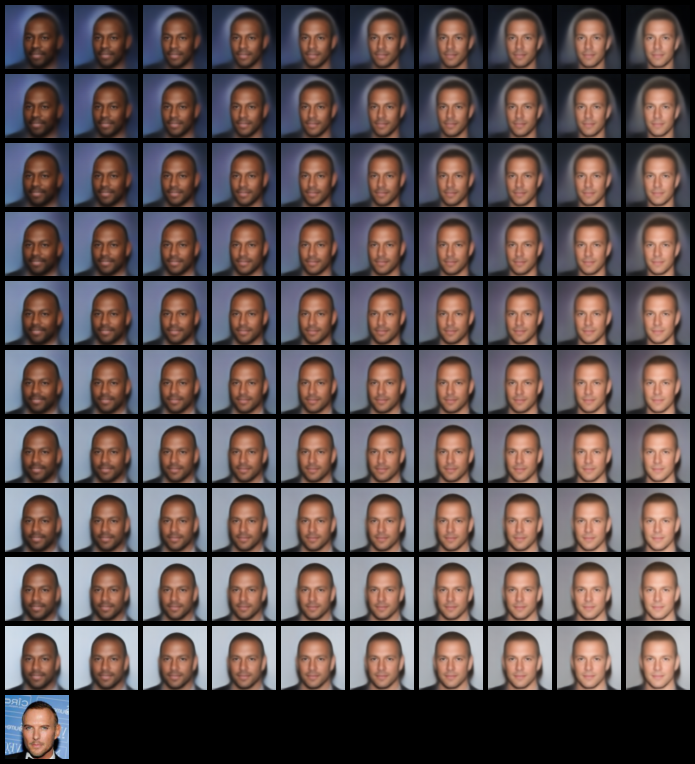}}
  \quad 
  \subcaptionbox{\label{fig:celeba2}}{\includegraphics[width=0.3\textwidth]{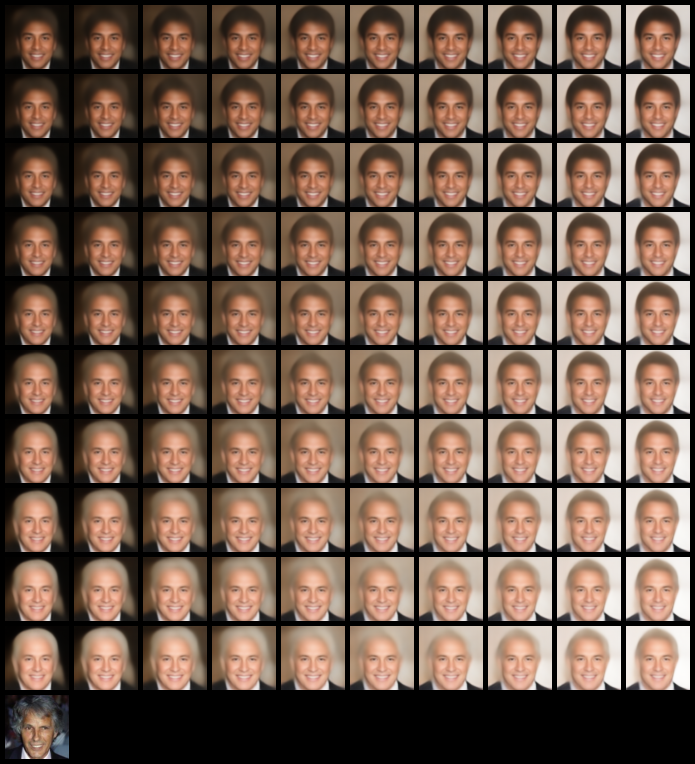}}
    \quad 
  \subcaptionbox{\label{fig:celeba3}}{\includegraphics[width=0.3\textwidth]{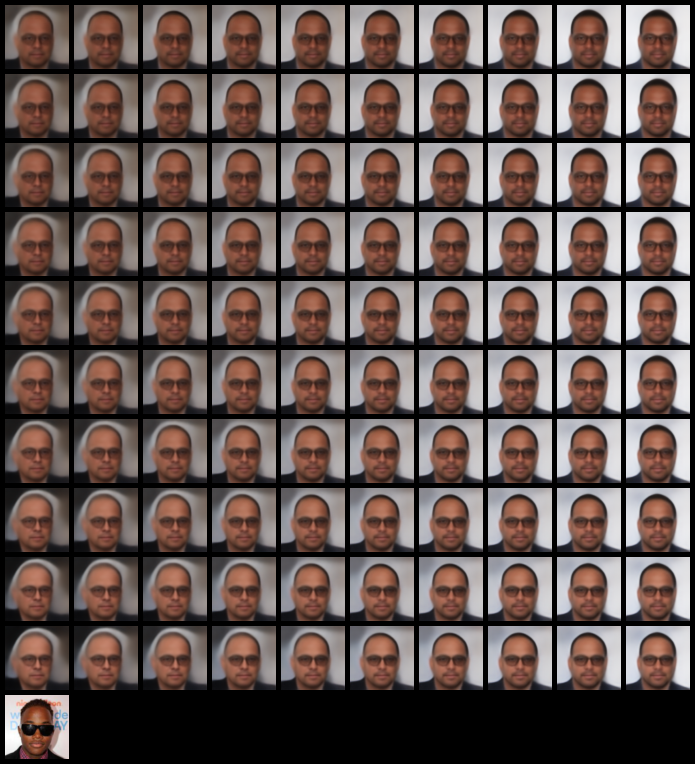}}
  \caption{Space of variations using a 2-dimensional free attribute on
    CelebA. The Variation Network is trained on representations
    obtained from a pre-trained VAE.
    Figures (b) and (c) are generated using the same model.}
  \label{fig:indbias}
\end{figure}

In order to check how crucial is the choice of the architecture for
the encoder and the decoder, we choose in this part to train a VarNet
on representations obtained from a pre-trained VAE, similarly to the
TwoStageVAE from \cite{dai2019diagnosing}. The first pre-trained VAE
has the same encoder as the one described in
Sect.~\ref{sec:impl-details} and a transposed VarNet-50 decoder. For
the encoder and decoder of the VarNet trained on these
representations, we use simple three-layered MLPs. The attribute
function network is, as in Sect.~\ref{sec:impl-details}, a
VarNet-50. The results in the case of a 2-dimensional free attribute
are displayed in Fig.~\ref{fig:indbias}. One striking observation is
the difference with Fig.~\ref{fig:celebafree} where changing the free
attributes corresponded to changing the background color. Such
difference is not surprising \cite{locatello2018challenging} and both
cases could be useful: removing information about background color in
$z$ could be of help for some downstream tasks where this piece of
information is irrelevant.
Here, in the case of   \ref{fig:indbias}, the free attributes tend to learn the ``pose'' and ``skin
color'' high-level attributes. This seems to be consistent across
different runs (e.g. Fig.~\ref{fig:celeba1} and \ref{fig:celeba2})
and consistent on different inputs (e.g. Fig.~\ref{fig:celeba2} and \ref{fig:celeba3}).

\section{Related work}
\label{sec:related-work}
The Variation Network generalizes many existing models used for
controlled input
manipulation by providing a unified probabilistic framework for this
task. We now review the related literature and discuss the connections
with VarNet.

The problem of controlled input manipulation has been considered in the \emph{Fader networks} paper
\cite{DBLP:journals/corr/LampleZUBDR17}, where the authors are able to
modify in a continuous manner the attributes of an input
image. Similar to us, this
approach uses an encoder-decoder architecture together with an
adversarial loss used to decouple templates and attributes. The
major difference with VarNet is that this model has a deterministic
encoder which limits the sampling possibilities as discussed in
Sect.~\ref{sec:underst-free-attr}. Also, this approach can only deal
with fixed attributes while VarNet is able to also learn meaningful free
attributes. In fact, VAEs \cite{2013arXiv1312.6114K}, WAEs
\cite{2017arXiv171101558T,2018arXiv180203761R} and Fader networks can
be seen as special cases of VarNet.

Recently, the \emph{Style Tokens} paper
\cite{DBLP:journals/corr/abs-1803-09017} proposed a solution to learn
relevant free attributes in the context of text-to-speech. The
similarities with our approach is that the authors condition  an
encoder model on an adaptive basis of style tokens (what we called
attribute space in this work). VarNet borrows
this idea but cast it in a probabilistic framework, where a
distribution over the attribute space is imposed and where the encoder
is stochastic. Our approach also allows to take into account fixed
attributes, which we saw can help shaping the free attributes.

Traditional ways to explore the latent space of VAEs is by doing
linear (or spherical \cite{white2016sampling}) interpolations between
two points. However, there are two major caveats in this approach: the requirement of always needing two points in order to
explore the latent space is cumbersome and the interpolation scheme
is arbitrary and 
bears no probabilistic interpretation.
Concerning the first point, a common approach is to find, a
posteriori, 
directions in the latent space that accounts for a particular change
of the (fixed) attributes \cite{upchurch2016deep}. These directions
are then used to move in the latent space.  Similarly,
\cite{DBLP:journals/corr/HadjeresNP17} proposes a model
where these directions of interest are given a priori. Concerning the
second point, \cite{laine2018feature-based} proposes to compute
interpolation paths minimizing some energy functional which result in
interpolation curves rather than interpolation straight
lines. However, this interpolation scheme is computationally demanding
since an optimization problem must be solved for each point of the interpolation path.


Another trend in controlled input manipulation is to make a
posteriori analysis on a trained generative model
\cite{DBLP:journals/corr/abs-1711-05772,pmlr-v80-adel18a,upchurch2016deep,2018arXiv180901859C}
using different means. One possible  advantage of these methods compared to ours
is that different attribute manipulations can be devised after the
training of the generative model. But, these procedures are still
costly and so provide any real-time applications where a user could
provide on-the-fly the attributes they would like to modify.
One of these approaches \cite{2018arXiv180901859C} consists in  using the trained
decoder to obtain a mapping $\mathcal{Z} \mapsto \mathcal{X}$ and
then performing
 gradient descent on an objective which accounts for the constraints
 or change of the attributes.
Another related approach proposed in
\cite{DBLP:journals/corr/abs-1711-05772} consists in training a
Generative Adversarial Network which learns to move in the vicinity of
a given point in the latent
space so that the decoded output enforces some constraints. The major
difference of these two approaches with our work is that these movements are done in a unique
latent space, while in our case we consider separate latent
spaces. But more importantly, these approaches implicitly consider
that the variation of interest lies in a neighborhood of the provided input.
In \cite{pmlr-v80-adel18a} the authors
introduce an additional latent space called \emph{interpretable lens}
used to interpret the latent space of a generative model. This space
shares similarity with our VarNet trained on VAE representations. The authors also
propose a joint optimization for their model, where the
encoder-decoder architecture and the interpretable lens are learned
jointly. The difference with our approach is that the authors optimize
an ``interpretability'' loss which requires labels and still need to
perform a posteriori analysis to find relevant directions in the
latent space.

\section{Conclusion and future work}
\label{sec:future-work}
We presented the Variation Network, a generative model able to vary
attributes of a given input. The novelty is that these attributes can
be fixed or learned and have a sound probabilistic
interpretation. 
Many sampling schemes have been presented together
with a detailed discussion and examples. We hope that the flexibility in the design
of the attribute function and the simplicity, from an implementation
point of view,  in transforming existing
encoder-decoder architectures  (it
suffices to provide the encoder and decoder networks) will be of
interest in many applications.

We saw that our architecture is indeed capable of decoupling templates from
learned attributes and that we have three ways of controlling the
free attributes that are learned: by modifying the hyperparameter terms in the
objective Eq.~(\ref{eq:lenc}), by carefully devising the attribute
functions or by working on different input representations with
different encoder/decoder architectures. 

For future work, we would like to extend our approach in two different ways:
being able to deal with partially-given fixed attributes and handling
discrete free attributes as in \cite{jeong2019learning}. We also want to investigate the use of 
stochastic attribute functions $\phi$. Indeed, it appeared to us that
using deterministic attribute functions was crucial and we would like
to go deeper in the understanding of the interplay between 
all VarNet components.



\bibliographystyle{abbrv}
\bibliography{varnet}

\end{document}